\documentclass{article}

\usepackage{arxiv}

\usepackage[utf8]{inputenc} 
\usepackage[T1]{fontenc}    
\usepackage{hyperref}       
\usepackage{url}            
\usepackage{booktabs}       
\usepackage{amsfonts}       
\usepackage{nicefrac}       
\usepackage{microtype}      
\usepackage{lipsum}		
\usepackage{graphicx}
\usepackage{natbib}
\usepackage{doi}
\usepackage{cite}
\usepackage{amsmath,amssymb,amsfonts}
\usepackage{algorithmic}
\usepackage{graphicx}
\usepackage{textcomp}
\usepackage{xcolor}
\usepackage{caption}
\usepackage{subcaption}
\usepackage{amsmath}
\usepackage{algorithm,algorithmic}
\usepackage{tabularray}

\title{Missing Data Imputation With Granular Semantics and AI-driven Pipeline for Bankruptcy Prediction}

\author{ \href{https://orcid.org/my-orcid?orcid=0000-0002-3131-012X}{\includegraphics[scale=0.06]{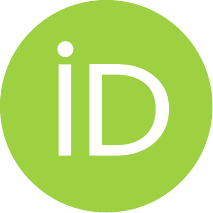}\hspace{1mm}Debarati B. ~ Chakraborty}\thanks{Use footnote for providing further
		information about author }\\
	School of Computer Science\\
	University of Hull\\
	Kingston Upon Hull, UK, HU6 7RX \\
	\texttt{debarati.earth@gmail.com} \\
	\And 
	{Ravi Ranjan} \\
	AMD India Private Limited\\
	11, Raheja Mindspace\\
	Hyderabad, India 500081\\
	\texttt{raviranjaniitj21@gmail.com}} 

\renewcommand{\shorttitle}{\textit{arXiv} Template}

\hypersetup{
pdftitle={A template for the arxiv style},
pdfsubject={q-bio.NC, q-bio.QM},
pdfauthor={David S.~Hippocampus, Elias D.~Striatum},
pdfkeywords={First keyword, Second keyword, More},
}

\begin{document}
\maketitle

\begin{abstract}
	This work focuses on designing a pipeline for the prediction of bankruptcy. The presence of missing values, high dimensional data, and highly class-imbalance databases are the major challenges in the said task. A new method for missing data imputation with granular semantics has been introduced here. The merits of granular computing have been explored here to define this method. The missing values have been predicted using the feature semantics and reliable observations in a low-dimensional space, that is, in the granular space. The granules are formed around every missing entry, considering a few of the highly correlated features to that of the missing value. A small set of the most reliable closest observations is used in granule formation to preserve the relevance and reliability, that is, the context, of the database against the missing entries within those small granules. An intergranular prediction is then carried out for the imputation within those contextual granules. That is, the contextual granules enable a small relevant fraction of the huge database to be used for imputation and overcome the need to access the entire database repetitively for each missing value. This method is then implemented and tested for the prediction of bankruptcy with the Polish Bankruptcy dataset. It provides an efficient solution for large and high-dimensional datasets even with large imputation rates. Then an AI-driven pipeline for bankruptcy prediction has been designed using the proposed granular semantic-based data filling method. The other two issues, i.e., high dimensional dataset, and high class-imbalance in the dataset have also been taken care of in this pipeline. The rest of the pipeline consists of feature selection with the random forest method to reduce the dimensionality, data balancing with synthetic minority oversampling (SMOTE), and prediction with six different popular classifiers including deep neural network. All methods defined here have been experimentally verified with suitable comparative studies and proven to be effective on all the data sets captured over the five years.  
\end{abstract}


\section{Introduction}

Bankruptcy, that is, the likelihood of failure of a company, is a major challenge in the financial sector. An average of 32,176 bankruptcies have been surveyed in the year between 2012 and 2016 only in the US\citep{chow2018analysis}. In all European countries, more than 2,00,000 companies file bankruptcy every year. Therefore, advance prediction of a company's bankruptcy would reduce the financial risk associated with the investors. The problem of bankruptcy prediction has been studied for decades and different solutions have been designed with different mathematical and statistical models to address this issue, but none of them seems to be very accurate. Nowadays a huge application of machine learning (ML) and artificial intelligence (AI) could be observed to address different challenges in the financial sector. Different ML and AI-based methods were designed to address the issues like credit risk assessment \citep{zakaryazad2016profit}, fraud detection in supply chain finance \citep{Rajagopal_23}, financial risk prediction \citep{Mashrur_20}, and prediction of investment risk \citep{Sun_22} etc. Different business sectors have already started using AI as a tool to enhance their businesses \citep{Qu_19}. Here in this work, we developed an AI-based solution for bankruptcy prediction.

The major underlying challenges that the financial data mostly encounters which make the deployments of AI/ ML models difficult could be summarized as follows \citep{leo2019machine}. i) Presence of missing entries in the large database, ii) high dimensionality, and iii) highly imbalanced training data. In the proposed work, the solution is designed in two stages. First, a new method of missing data imputation has been defined with granular semantics, which makes the imputation in the big bankruptcy data computationally less expensive, and an AI-driven pipeline is followed for predicting bankruptcy by addressing the aforementioned challenges. 

Missing values is a major challenge in data quality. It is a real-life issue to deal with. There are several reasons for these missing entries in the data sets. These include errors in data collections or data entries, unavailability of the required information, incomplete features, and incomplete information, etc. \citep{HASAN2021100799}. As there is no alternative to the prediction of the missing values, different ML-based and statistical models have been designed so far to address this issue \citep{Alabadla_22}. The method defined here explores the merits of granular computing to judiciously deal with issues like large size and high dimensionality associated with the bankruptcy database. 

Granulation is a basic step in the human cognition system and therefore a part of natural computation \citep{DChakraborty_21}. According to the concept, as introduced by Zadeh \citep{Zadeh_97}, granulation involves the partition of an object into granules, a granule being a group of elements drawn together by indistinguishability, equivalence, similarity, and functionality. Granular computing has been used to address different problems in data science, including large-scale group decision-making \citep{Zheng_22}, video analysis \citep{DChakraborty_21, Chakraborty_23}, time series prediction \citep{Ma_22}, fire threat prediction \citep{CHAKRABORTY2022102140}, etc. Formation of granules and computation with granules are the two primary phases in granular computing, and those vary based on the applications. That is, how to draw the group of elements together in a dataset, and what to do with the small amount of information depends on the problems to be solved. In this work, the aim is to predict missing values using a small amount of relevant information in the database. The bankruptcy datasets are usually high-dimensional and contain tens of thousands of observations, that is, big-sized data that may have thousands of missing elements. To reduce the imputation complexity caused by these large datasets, the granules are formed around the missing values, considering only the most semantic features of the entries. A few observations close to the said entry and located over those correlated features are used to form the granules. The granules would not consider any other missing entries, except the seed point (the point around which it is formed). This is how both the relevance and reliability of the large dataset around the missing values are being preserved in the small fraction of the database. Intergranular predictions are performed here within the semantic granules for imputation which pretermit the need to access the entire huge database again and again for every missing entry, thereby reducing the computational complexity without affecting the accuracy even with an increasing amount of missing entries.      

\begin{figure}[h]
    \centering
    \includegraphics[scale=0.6]{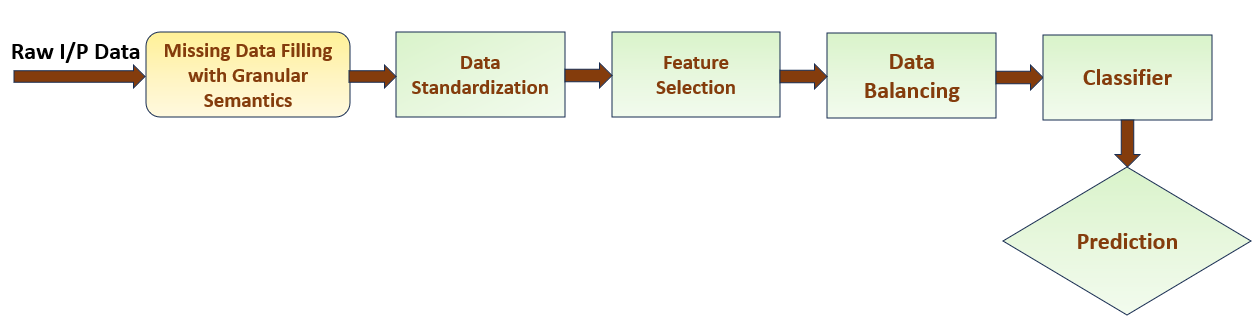}
    \caption{Pipeline for bankruptcy prediction}
    \label{BlockDia}
\end{figure}

Once all missing values are predicted in the given challenge of bankruptcy prediction, the remaining steps should be taken to achieve the goal. Here, we have defined the end-to-end pipeline as shown in Fig. \ref{BlockDia}. As could be observed in Fig. \ref{BlockDia}, data filling would be followed by feature selection with the random forest method \citep{Paul_18} to reduce high dimensionality. Then data balancing would be performed with synthetic minority oversampling technique (SMOTE) \citep{chawla2002smote} since the number of bankrupted companies is much less than those of non-bankrupted ones in a dataset, and this imbalance could adversely affect the ML-based classifiers. The pipeline was then tested with six different standard classifiers, including a deep neural network, and it was proven to be effective in prediction. 

The novelties of the proposed work could be summarized as follows. i) Defining a new method for missing data imputation with reduced complexity, ii) formulating contextual granules by preserving the relevance and reliability of the database in its small fraction against the missing entries, and iii) designing a pipeline for bankruptcy prediction by addressing the other challenges like multidimensionality and data imbalance.  

The remainder of the article is organized as follows. Sec. \ref{LS} describes a few relevant works on missing data imputation and bankruptcy prediction. Sec. \ref{PropMeth} contains the method defined here for missing data filling. The theoretical details on the formation of contextual granules, granular imputation, and the stepwise algorithm to predict missing values are explained in this section. The pipeline, designed here for bankruptcy prediction is elaborated in Sec. \ref{method}. All the theories defined here are validated with experimental outcomes and suitable comparative studies in Sec. \ref{exp}. The overall conclusion of the article is drawn in Sec. \ref{con}.

\section{Related Work}\label{LS}
\subsection{Missing Data Imputation}
The problem of missing data in a dataset has been addressed widely since the last couple of decades. Here we are going to discuss only a few of the benchmark methods. A very popular approach to this problem was filling up the missing entries with some constants, like zeros or the mean of the distribution \citep{Little_19}. Hastie Yan \emph{et al.} \citep{Hastie_99} first introduced a method for missing value imputation with the k-nearest neighbor, but it was not very effective with a high imputation rate. Yan \emph{et al.} introduced an approach of missing data filling with Gaussian mixture model in \citep{Yan_15}, where the imputation was carried out iteratively from the clusters satisfying log likelihood, but this method failed to satisfy class likelihood since there was a huge shared common region between the classes.   

\subsection{Bankruptcy Prediction}
In the early years, the methods of linear regression, discriminant analysis, and logistic regression were used for bankruptcy classification tasks. The most widely accepted and used subset of machine learning models for predicting financial distress is illustrated in the following paragraph.

 Machine Learning models were deployed by many researchers, including \citep{leo2019machine}, for banking risk management. The authors of \citep{mai2019deep} use deep learning models for the same purpose. Other authors \citep{smiti2020bankruptcy} use deep learning for borderline smote, wherein they focused on imperfect classification. The authors in \citep{zikeba2016ensemble} use ensemble-boosted trees for bankruptcy prediction. Similar work has been done by the authors in \citep{zakaryazad2016profit} for fraud detection and direct marketing using Artificial Neural Networks. The authors in \citep{wang2017bankruptcy} use autoencoder techniques, and neural networks with dropouts and compare the existing proposed models. The authors in \citep{aniceto2020machine} use Logistic regression as the benchmark model for comparing the results of different machine-learning techniques. This model can be used for classification tasks wherein it is used to describe a data relationship between dependent and independent variables. It performs predictive analysis. The authors of \citep{chen2016financial} use the k-nearest neighbors algorithm (k-NN) as a machine learning method without parameters for the classification of Bankruptcy vs. Non-Bankruptcy and achieved a decent score. and achieved good classification accuracy\citep{filletti2020using}. The authors in \citep{leo2019machine} use a Decision Tree as a classifier for better prediction by allocating weights to it, making decisions that are easy to infer. It is considered a non-parametric algorithm due to the tree size growing to match the classification problem's complexity\citep{bellovary2007review}. Here the most relevant feature acts as the root node, and the following relevant features form its child. The authors of \citep{zikeba2016ensemble} use multiple decision trees in a combined form to represent random forests or random decision forests, an ensemble learning method used for classification and regression tasks as training and outputs the class based on classification, and found very high accuracy. The authors of \citep{pawelek2019extreme} and \citep{kumar2007bankruptcy} use a gradient-boosting algorithm to predict the bankruptcy of Polish Companies. Firstly, it is used to remove the outliers from the dataset and then to predict bankruptcy. In this paper, the authors indicated that by removing the outliers from the dataset using gradient boosting, it is possible to increase the prediction rate. The authors of \citep{mai2019deep} use Neural networks to predict the accuracy and found that it outperforms the accuracy as compared with all existing machine learning models. Like each neuron in our brain comes up with a simple task and controls the complex and challenging functions, cognitive tasks, etc.\citep{jouzbarkand2013creation} Using logistic regression, each neuron can be related mathematically, and therefore the overall artificial neural network can be considered as multiple logistic regression classifiers attached to each other. \citep{mai2019deep}

\section{Proposed Methodology on Missing Data Filling}\label{PropMeth}
\label{sec:headings}
Let $\Phi$ be a data set in $d$ dimensional feature space $\mathbb{F}^d$. Let there be $N$ numbers of observations listed in the dataset. Therefore, each characteristic vector $\overline{\mathbb{f}^i} \in \mathbb{F}^d$ is supposed to have dimension $N\times 1$. $\Phi$ is supposed to contain $N\times d$ number of data points, but among which several of them are missing. The models, that we got trained with $\Phi$ won't be very reliable in the presence of these missing values. Here in this, we have formulated a new method of missing value prediction considering the feature semantics and inter-granular distribution. The entire work can be subdivided into three segments, viz. i) finding missing values and conversion of categorical values to numerics, ii) computation of feature semantics and formation of contextual granules, and iii) granular imputation of missing values. 

\subsection{Categorical to Numerical Conversion}\label{c2nsec}

In this work, we assumed all features to be numerical. But in practice, the presence of categorical features is as relevant as the numerical ones. We are going to address this issue now. We convert categorical values to numeric values within the range of $0-1$. Let a feature vector, $\overline{f}$ be categorical. Let $|f|=C$, where $|.|$ represents the cardinality of a vector. Let the possible values of the elements in $\overline{f}$ be contained in the tuple $\{v_1,v_2,...,v_C\}$, where all $v_1, v_2,...,v_C$ are categorical. The categorical to numerical conversion of this tuple is done following Eqn. \ref{c2n} where $f_c$ represents the categorical values of the elements and $f'_c$ represents the numerical values after conversion to the feature vector $\overline{f}$.

\begin{equation}
f'_c=
\begin{cases}
  \frac{1}{c}, & \text{if }
       \begin{aligned}[t]
       f_c=v_1
       \end{aligned}
\\
\frac{2}{c}, & \text{if }
       \begin{aligned}[t]
       f_c=v_2
       \end{aligned}
\\
.\\
.\\
.\\
1 & \text{if }
       \begin{aligned}[t]
       f_c=v_C
       \end{aligned}
\end{cases}\label{c2n}
\end{equation}

Finding missing values in the given data set is an underlying challenge that we need to address. In this work, we have used the method as described by Kachuee \emph{et al.} in \citep{Kachuee22}. A mask vector $K$, where $k_j \in \{0,1\} \forall k_j \in K$ is used to identify the missing data points. The missing values in a dataset are generally represented by 'NaN' or '?'. The values of the elements in the mask vector $K$ is determined according to Eqn. \ref{kavl}. $\Phi_j$ represents the elements present in the dataset $\Phi$ in Eqn. \ref{kavl}.

\begin{equation}
k_j=
\begin{cases}
  0, & \text{if }
       \begin{aligned}[t]
       \Phi_j&=\varnothing
       \end{aligned}
\\
  1, & \text{otherwise}
\end{cases}\label{kavl}
\end{equation}

\subsection{Formation of Contextual Granules}\label{FCGr}
The formation of granules and the computation with the granules are the two primary components of granular computing. Given the fact that granularity is a basic component of the human cognition system, the aspect of granularity is equally important while heading toward a specific solution. Granules could be formed considering different types of similarities in a data set. It could be value-based similarities, spatial similarities, distributional similarities, and many more. Granules play a vital role in this work because the prediction of a missing value is completely dependent on the formation of a granule. To make the prediction more accurate, a way to form granules with closely related features has been developed here. Since we are dealing here with multi-dimensional data, the granules are to be formed in the nearby dimensions. 

The feature correlation is measured here with the Pearson Coefficient. That is, let $\overline{\mathbb{f}^x}$ and $\overline{\mathbb{f}^y}$ be two feature vectors in the feature space $\mathbb{F}^d$. Each feature vector contains $N$ observations. The similarity between $\overline{\mathbb{f}^x}$ and $\overline{\mathbb{f}^y}$, $\rho_{x,y}$ is measured as per Eqn. \ref{rho}. In Eqn. \ref{rho} $\sigma_{x,y}$ represents the covariance between $\overline{\mathbb{f}^x}$ and $\overline{\mathbb{f}^y}$, while $\sigma_x$ and $\sigma_y$ denote the standard deviations of $\overline{\mathbb{f}^x}$ and $\overline{\mathbb{f}^y}$ respectively.
\begin{equation}\label{rho}
    \rho_{x,y}=\frac{\sigma_{x,y}}{\sigma_x\sigma_y}
\end{equation}

\begin{figure}[h]
    \centering
    \includegraphics[scale=0.5]{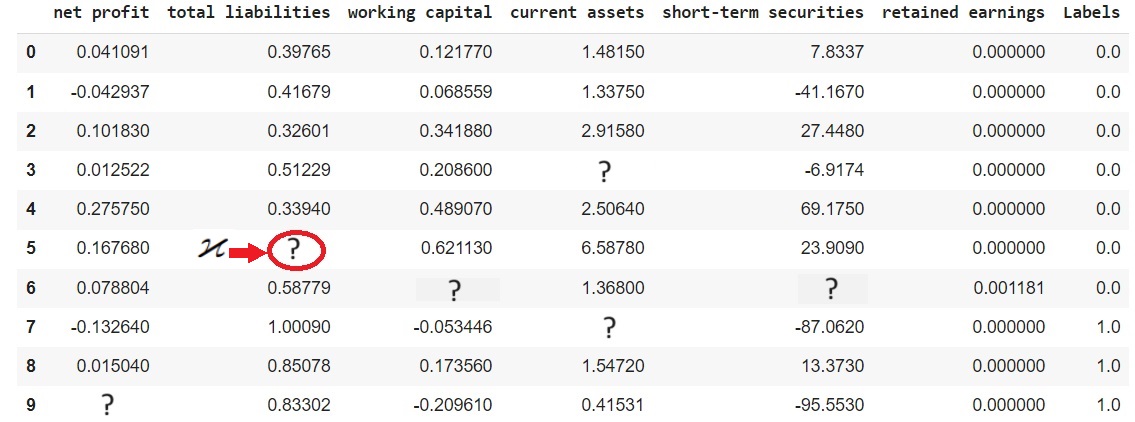}
    \caption{An example data-set with missing entries}
    \label{ExData}
\end{figure}

\begin{figure}[h]
    \centering
    \includegraphics[scale=0.5]{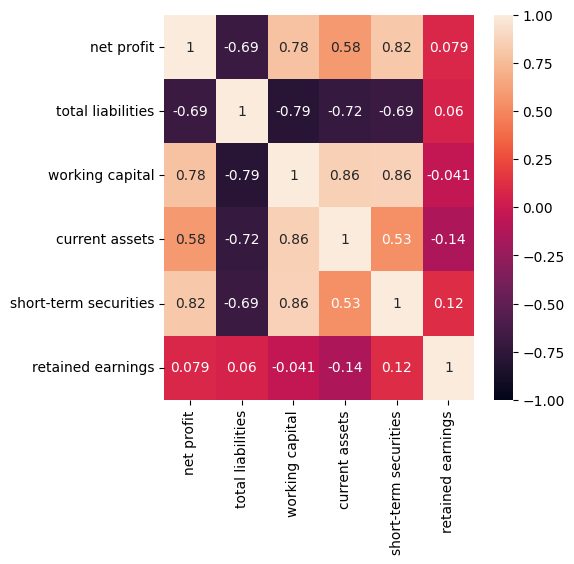}
    \caption{Input Feature Correlation Matrix of the data-set in Fig. \ref{ExData}}
    \label{ExDataCorr}
\end{figure}

To provide further clarity in the concept, an example data set with missing values and its corresponding correlation matrix are shown in Figs. \ref{ExData} and \ref{ExDataCorr}, respectively. This example dataset Fig. \ref{ExData} contains six input features, while the seventh column represents the output label. The entries '?' represent the missing values in the data set.

In the data set $\Phi$, with $d$ feature vectors in feature space $\mathbb{F}^d$, $d\times d$ correlation matrix ($\Gamma$) would be generated. Let the granule formed around the missing values be of dimension $\delta$, where $\delta<<d$. Let $\varkappa$ be a missing element in the data set $\Phi$. Let the location of $\varkappa$ in $\Phi$ be $(\alpha,\beta)$. Assuming that the feature vectors are column-wise populated in $\Phi$, it can be stated that $\varkappa\in \overline{\mathbb{f}^\beta}$. Since the work focuses on considering the semantics to predict the missing values, the closest $\delta$ number of features around $\overline{\mathbb{f}^\beta}$ should be identified first. In this work, the feature semantics is taken into account using Eqn. \ref{fd}. In Eqn. \ref{fd} $\Gamma^\beta$ represents the $\beta^{th}$ row of the correlation matrix $\Gamma$, and $\rho_{\beta,i}$ denotes the similarity between the features $\overline{\mathbb{f}^\beta}$ and $\overline{\mathbb{f}^i}$ according to Eqn \ref{rho}, where $i\in \mathbb{F}^d$. Finally, $\Gamma^\beta_\delta$ in Eqn. \ref{fd} contains the set of $\delta$ number of features with maximum correlation to $\overline{\mathbb{f}^\beta}$. 
\begin{equation}\label{fd}
\begin{split}
    \Gamma^\beta_\delta=max\{\delta\}(\Gamma^\beta)&=\{\rho_{\beta,1}, \rho_{\beta,2},..., \rho_{\beta,\delta}\}\\ &:\rho_{\beta,i}=max(\Gamma^\beta|\{\rho_{\beta,i+1}, \rho_{\beta,i+2},..., \rho_{\beta,d}\})
\end{split}    
\end{equation}

Once the semantics of the granule is determined, the according observations need to be considered to form the granule. Let $\eta$ number of entries be selected around $\varkappa$, that is, around the $\alpha^{th}$ observation, where $\eta<<N$. The missing values in the observed space should be avoided to ensure the reliability of the granule. Therefore, the set of observations to be considered from the data set $\Phi$ is determined by Eqn. \ref{rval}. $\Upsilon$ represents the set of $\alpha$ number of observations to be considered. It is clear from Eqn. \ref{rval}, that if any observation $(\alpha-i)$ contains a missing value, that entry will be replaced with $(\alpha-\eta-i)^{th}$ one where $(\alpha-\eta-i)\in N$.

\begin{equation}
\Upsilon=
\begin{cases}
  \bigcup n : n=\alpha-\eta,...,\alpha-i,...,\alpha-1, & \text{if }
       \begin{aligned}[t]
       \Phi(n,m)&\neq\varnothing: \forall ~ m \in \Gamma^\beta_\delta
       \end{aligned}
\\
  \bigcup n: n=\alpha-\eta,...,\alpha-\eta-i,...,\alpha-1, & \text{if }
       \begin{aligned}[t]
       \Phi(\alpha-i,m)&=\varnothing: \exists ~ m \in \Gamma^\beta_\delta
       \end{aligned}
\end{cases}\label{rval}
\end{equation}

In order to relate the mathematical formulation to the example dataset of Fig. \ref{ExData}, the query point $\varkappa$ has been shown with a red circle there. According to the example data set, $\beta=total ~ liabilities$. It could be stated from the correlation matrix of Fig. \ref{ExDataCorr} and Eqn. \ref{fd} that $\Gamma^\beta_\delta=\{working ~ capital, ~ current ~ assets\}$ if $\delta=2$. In the same way, according to Fig. \ref{ExData}, $\alpha=6$. Following Eqn. \ref{rval}, $\Upsilon=\{5,3\}$ if $\eta=2$ since the fourth row contains a null value against the feature $current ~ assets\in \Gamma^\beta_\delta$.  

The semantic granule around the point $\varkappa$, $\gamma_\varkappa$ is now formed as per Eqn. \ref{gr}. It can be claimed that these granules will always preserve the context of the missing data, and are also reliable. It is because the granule contains only the information from $\Gamma^\beta_\delta$ to retain the semantics, and from $\Upsilon$ to assure the reliability. 
\begin{equation}\label{gr}
   \gamma_\varkappa=\bigcup\Phi_{x,y} ~ \forall ~ (x\in \Upsilon ~ \& ~ y\in\Gamma^\beta_\delta) 
\end{equation}

\subsection{Granular Imputation of Missing Values}\label{imputation}
Once the granules are formed following the process described in Sec. \ref{FCGr}, the estimation of the missing values are to be done with these granules. That is, the value of the point $\varkappa$ should be done using the information contained there in the granule $\gamma_\varkappa$ (see Eqn. \ref{gr}). Here a linear regression model is used for prediction. The difference between the granular prediction and conventional linear regression could be summarized as follows. 

\begin{itemize}
    \item The observation space has been reduced from a dimension of $N \times d$ to $\eta \times \delta$, where $\eta<<N$ and $\delta<<D$. 
    \item Only the features contained in $\Gamma^\beta_\delta$ (see Eqn. \ref{fd}) would be considered as the input feature and $\overline{\mathbb{f}^\beta}$ would be considered as the output feature, to ensure that you have the best possible line of estimation for the particular missing entry in the given dataset.
    \item The observations contained in $\Upsilon$ (see Eqn. \ref{rval}) guarantee that no missing entries should be used to train the regressor, thus increasing the reliability of the regressor.
\end{itemize}

As stated earlier, the information contained in $\gamma_\varkappa$ will be used here for training and estimation. Similarly, the dimension of the training data set would be of $(\eta-1)\times \delta$, with $(\eta-1)\times (\delta-1)$ number of input values and $(\delta-1)$ number of their corresponding output values. Therefore, from Eqn. \ref{gr} the input training data ($X^T_\varkappa$) and its corresponding output values ($Y^T_\varkappa$ ) for $\varkappa$ could be written according to Eqn. \ref{TrainX} and Eqn. \ref{TrainY} respectively.
\begin{equation}\label{TrainX}
X^T_\varkappa =
\begin{bmatrix}
    \Phi_{\Upsilon(1),\Gamma^\beta_\delta(1)}       & \Phi_{\Upsilon(1),\Gamma^\beta_\delta(2)} & \dots & \Phi_{\Upsilon(1),\Gamma^\beta_\delta(\delta-1)} \\
    \Phi_{\Upsilon(2),\Gamma^\beta_\delta(1)}       & \Phi_{\Upsilon(2),\Gamma^\beta_\delta(2)} & \dots & \Phi_{\Upsilon(2),\Gamma^\beta_\delta(\delta-1)} \\
    \hdotsfor{4} \\
    \Phi_{\Upsilon(\eta-1),\Gamma^\beta_\delta(1)}       & \Phi_{\Upsilon(\eta-1),\Gamma^\beta_\delta(2)} & \dots & \Phi_{\Upsilon(\eta-1),\Gamma^\beta_\delta(\delta-1)} 
\end{bmatrix}
\end{equation}

\begin{equation}\label{TrainY}
Y^T_\varkappa =
\begin{bmatrix}
   \Phi_{\Upsilon(1),\beta} \\
    \Phi_{\Upsilon(2),\beta} \\
    \hdotsfor{1} \\
   \Phi_{\Upsilon(\eta-1),\beta} 
\end{bmatrix}
\end{equation}

Now, a multivariate regression model will be generated with $X^T_\varkappa$ and $Y^T_\varkappa$. Therefore, Eqn. \ref{regr} can be written to map the relation between $X^T_\varkappa$ and $Y^T_\varkappa$. In Eqn. \ref{regr} $\Theta=\{\theta_1, \theta_2,...,\theta_{delta-1}\}$ represent the coefficients and $\epsilon$ represent the error. 
\begin{equation}\label{regr}
    Y^T_\varkappa = X^T_\varkappa\cdot \Theta + \epsilon
\end{equation}

The values of $\Theta$ will be estimated with the least squares estimation; that is, $\hat{\Theta}$ would retain optimal values in the error plane, to minimize the error in estimation. Now, the missing value of $\Phi_{\alpha, \beta}$ would be estimated on the basis of the model. In this estimation, the input values will be $X^s_{\varkappa}=[\Phi_{\Upsilon(\eta),\Gamma^\beta_\delta(1)}, \Phi_{\Upsilon(\eta),\Gamma^\beta_\delta(2)},...,\Phi_{\Upsilon(\eta),\Gamma^\beta_\delta(\delta-1)}]$. Given the set of input, the missing value would be estimated using Eqn. \ref{est}.
\begin{equation}\label{est}
    \hat{\Phi}_{\alpha,\beta} = X^s_{\varkappa}\cdot \hat{\Theta}
\end{equation}

\begin{figure}[h]
    \centering
    \includegraphics[scale=0.4]{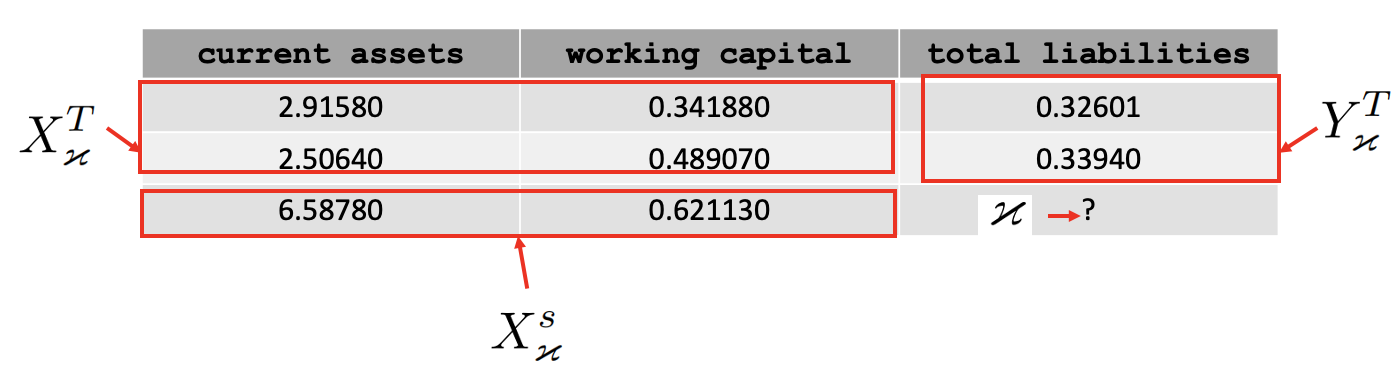}
    \caption{An example granule $\gamma_\varkappa$ around the missing value $\varkappa$}
    \label{ExGranule}
\end{figure}

Similarly, all missing values in the data set $\Phi$ would be imputed using the proposed granular estimation. An example granule around the missing point $\varkappa$ of is shown in Fig. \ref{ExGranule} against the data set given in Fig. \ref{ExData}. In the example shown there in Fig. \ref{ExGranule} $\delta=\eta=2$. 

\subsection{Algorithm for missing value prediction with granular semantics}

A stepwise description of the method for missing value prediction with granular semantics is summarized here as Algorithm 1. If the dataset $\Phi$ with missing values is provided as the input to the algorithm, the output dataset $\hat{\Phi}$ would have all those values imputed with granular semantic prediction. 

\begin{algorithm}
 \caption{Missing Value Prediction with Granular Semantics}          
\label{algSPO}                           
\begin{algorithmic}                    
    \STATE INPUT: Data set $\Phi$ with missing values
    \STATE OUTPUT: $\hat{\Phi}$ with filled in values
    \STATE INITIALIZE: Missing Values=$\varnothing$
    \item 1: Convert all the categorical features to numerical values using Eqn. \ref{c2n}
    \item 2: Find out the missing values in $\Phi$ with Eqn. \ref{kavl}
    \item 3: For a certain missing entry $\varkappa={\Phi}_{\alpha,\beta}$ at position $(\alpha, \beta)$ do the following
    \item 4: Find out the most similar $\delta$ number of features of $\beta$, $\Gamma^\beta_\delta$, using Eqn. \ref{fd} 
    \item 5: Find out the closest $\eta$ number of observations of the $\alpha^{th}$ one, without any missing entries, $\Upsilon$, using Eqn. \ref{rval}. 
    \item 6:Form the semantic granule $\gamma_\varkappa$ using Eqn. \ref{gr}.
    \item 7:Predict the missing value, $\hat{\Phi}_{\alpha,\beta}$ using a regression model within the granule $\gamma_\varkappa$ as described in Sec. \ref{imputation}.
    \item 8: Impute the value $\hat{\Phi}_{\alpha,\beta}$ in the estimated data set $\hat{\Phi}$
    \item 9: Repeat steps 4 to 8 for each missing value in $\Phi$
    \item 10: Output $\hat{\Phi}$
    \end{algorithmic}
\end{algorithm}

\section{Pipeline for Bankruptcy Prediction}\label{method}
The underlying objective of this work is to develop a pipeline for bankruptcy prediction. The proposed data pre-processing method has already been discussed in Sec. \ref{PropMeth}. Once all the missing values are filled in the data set $\Phi$, the next step would be to extract the relevant features for prediction. The datasets can be biased at times. High-class imbalance is a major issue in the bankruptcy data set as the number of bankrupted organizations is very few compared to the non-bankrupted ones. This phenomenon can induce a bias in the ML-based model, resulting in failure in prediction. {This can be overcome by uncorrelated sets of features and equal odds, i.e., choosing an equal number of true-positives and false-positives for each class. Many of the supervised and unsupervised methods show varying results depending on features taken for model building and specific datasets}. The proposed pipeline of the risk prediction model is shown in Fig. \ref{BlockDia}. Block-wise explanation of the model is provided in the following section.  The pipeline has been validates with six classifiers, like, Logistic Regression, Random Forest, Decision Tree, Gradient Boosting, K-Nearest Neighbor, and Artificial Neural Network.\\

\subsection{Data Standardization}
Each data point is standardized in such a way that all features have unit variance and zero mean. The standardization would take place using Eqn. \ref{stnd}. In Eqn. \ref{stnd}, $\Phi_{m,n}'$ represents the standardized value of the data point $\Phi_{m,n}$, $\overline{\mathbb{f}^n}'$ and $\sigma_n$ represent the mean and standard deviation respectively of the feature vector $\overline{\mathbb{f}^n}$.  
\begin{equation}\label{stnd}
    \Phi_{m,n}'=\frac{\Phi_{m,n}-\overline{\mathbb{f}^n}'}{\sigma_n}
\end{equation}

\subsection{Feature Selection}
Since the bankruptcy data used to be high dimensional, a feature reduction method has been used in the second phase of the pipeline to enhance the efficiency of the method. Here Random Forest \citep{Paul_18} has been included for this task. In Random Forest, each tree is a hierarchy of true or false questions based on a single feature or multiple features. The tree splits the dataset into two partitions at each node, similar observations are stored in one partition, and different observations from the first partitions are stored in another partition. That is why the importance of each feature depends on the purity of each partition.

When a tree is trained, the amount of reduction in an impurity of each feature can be calculated.  The feature which is reducing impurities is a more important feature. In random forests, the reduction in impurity by each feature can be averaged for all trees to know the importance of a feature. 
For better understanding, the features extracted or chosen at the top of the tree are more important than the bottom nodes, because the top node has a large amount of information or entropy.\\

\subsection{Data Balancing}
As discussed earlier, bankruptcy datasets are highly class-imbalanced. That is, the proportion of positive samples and negative samples in the training data is far away from equity. Here in this work, the Borderline  Synthetic Minority Over-Sampling Technique, i.e., SMOTE \citep{chawla2002smote} has been used to handle the issue. SMOTE performs an oversampling task, which means it increases the minority classes of data. The SMOTE technique has a unique feature. It does not duplicate observations. It produces new data points corresponding to features along with the randomly selected point and their nearest neighbors.\\

\subsection{Model Selection}\label{ModelSel}
In this work, we have validated the proposed pipeline with six different prediction models. These models are standard classification and prediction models in machine learning. The following list provides those 6 models that are trained and tested for bankruptcy prediction.
\begin{itemize}
    \item  Logistic Regression
    \item  K- Nearest Neighbor
    \item  Decision Trees
    \item  Random Forest
    \item  Gradient Boosting
    \item  Deep Neural Network
\end{itemize}

\section{Experimental Outcomes}\label{exp}

\subsection{Dataset}
Here in this work Polish companies bankruptcy datasets \citep{polish_Bankruptcy} have been used for experimentation. The data set is about the prediction of bankruptcy of Polish companies. This data set contains 64 quantitative features. This dataset describes the bankruptcy status of Polish companies. This dataset is generated from EMIS (Emerging Market Information Service) dataset. This data was collected within the time period of 2000 to 2013. It contains two classes: class 0 and class 1. Class 0 shows that the company is not bankrupt and class 1 shows the Polish bankrupt companies. The input features that are used in the dataset are net profit, total liabilities, working capital, current assets, retained earning, EBIT, book value of equity etc. (see \citep{polish_Bankruptcy} for details). The dataset contains the observations of five years, during 2007-2013 among which 7027 instances are given in the first year, 10173 in the second year, 10503 in the third year, 9792 in the fourth year, and 5910 in the fifth year. The dataset contains several missing values, and it is also highly imbalanced.

The number of missing entries present in the Polish Bankruptcy data set \citep{polish_Bankruptcy} has been listed here in Table \ref{MV}. Here in this work these missing values have been filled with the method described in Sec. \ref{PropMeth}. Please note, the proposed method has been implemented using Python 3.9 in Google Colab. 

\begin{table}[ht]
    \centering
    \caption{Missing Values in Polish Bankruptcy Data}
    \begin{tabular}{|c|c|c|}
    \hline
        Data & $Total~ data~ points$ & $Missing~ entries$ \\
         \hline
        $1^{st} Year$ & $7027\times 64$ & $5838$  \\
        \hline
$2^{nd} Year$ & $10173\times 64$ & $12157$  \\
        \hline
$3^{rd} Year$ & $10503\times 64$ & $9888$  \\
        \hline
$4^{th} Year$ & $9792\times 64$ & $8777$  \\
        \hline
$5^{th} Year$ & $5910\times 64$ & $4666$  \\
        \hline
    \end{tabular}
    \label{MV}
\end{table}

\subsection{Effectiveness of Granular Semantics-based Missing Value Prediction}

This section experimentally demonstrates the effectiveness of missing value prediction with the proposed granular semantics-based data-filling method (GS) which is described in Sec. \ref{PropMeth}. In order to prove the utility of the method, a comparative study has been performed with four other benchmark data imputation methods. Here one recent method, from each of the four standard processes of data imputation has been selected for the sake of comparative studies. Those methods are listed as follows. 

\begin{itemize}
    \item MICE: Multivariate Imputation by Chained Equation \citep{Azur_11} uses multiple iteration for missing data imputation. In \citep{Azur_11} linear regressor has been used iteratively as a predictive model to fill in all the missing values.
    \item Fractional Hot Deck Imputation (FHDI): Here in this work \citep{Song_20} each missing value has been replaced with a set of weighted imputed values here a missing value of the recipient unit gets replaced by the similar values of the donor unit. The values of donor unit are assigned with fractional weights in this prediction. 
    \item Autoencoder: Autoencoders have become popular now-a-days for missing value imputation \citep{Gjorshoska_22}. Here the autoencoder approximates the values by learning a higher-level representation of its input. 
\end{itemize}

The comparative studies among these aforementioned methods and the proposed granular semantic-based one have been performed here. Synthetic imputation has been performed on the data-set, that is, some random real values have been replaced with null, and different imputation methods are performed in order to check how close the predictions are. The closeness, or the error in prediction has been measured using Eqn. \ref{measure}. That is the normalized error between the predicted ($\hat{\Phi}_{m,n}$) and real value ($\Phi_{m,n}$) has been computed over the feature $\overline{\mathbb{f}_n}$. 

\begin{equation}\label{measure}
    Error=\frac{|\Phi_{m,n}-\hat{\Phi}_{m,n}|}{|max(\overline{\mathbb{f}_n})-min(\overline{\mathbb{f}_n)}|}
\end{equation}

\begin{figure}[h]
    \centering
    \includegraphics[scale=0.6]{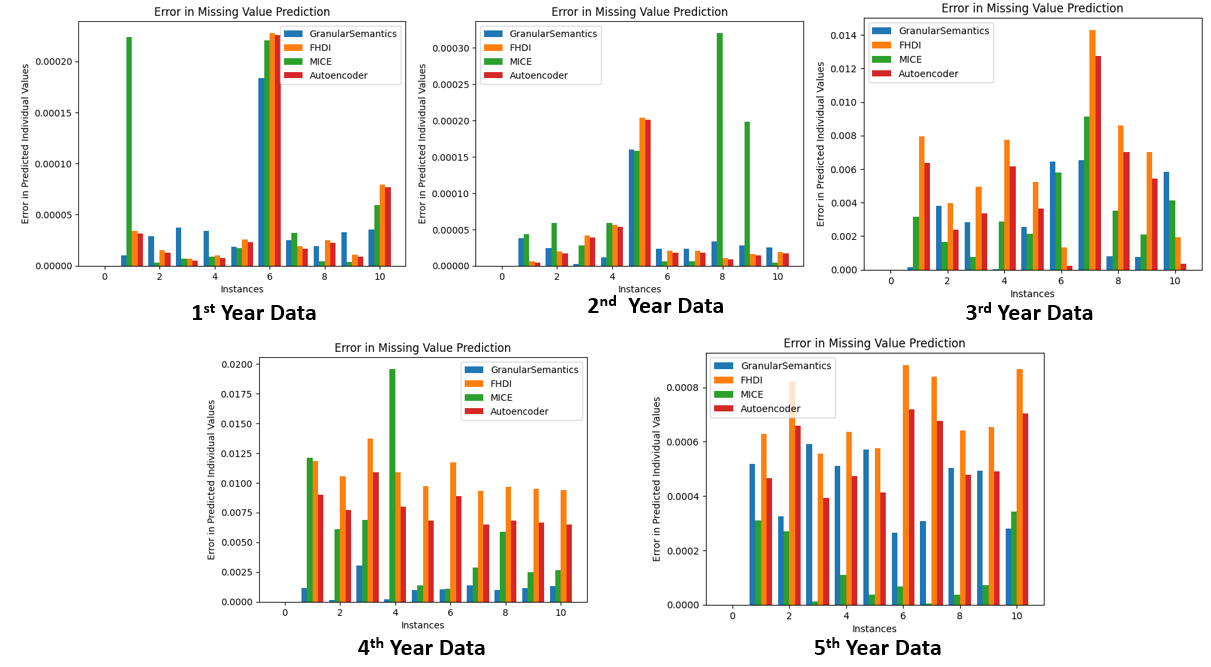}
    \caption{Error in individual value prediction for year-wise Polish Bankruptcy Data}
    \label{IE}
\end{figure}

Fig. \ref{IE} shows the error in the prediction of individual values with different methods. As it could be observed from the figure the proposed granular prediction method results in low error consistently over all the years. The performance of FHDI and Autoencoder are equally good in most of the cases. Please note that FHDI needs to repeat the regression process several times with different weights for a single value. On the other hand, the autoencoder must develop an encoder and decoder architecture with high-level representation learning. Compared to the aforesaid methods, the proposed method would produce a prediction only with a very small segment of the dataset, in these experiments we considered $\delta=5<<d=64~ and ~ \eta=7 << N\cong 10,000 $, and with a single regression for a value by exploring the merits of granulation.   

\begin{figure}[h]
    \centering
    \includegraphics[scale=0.5]{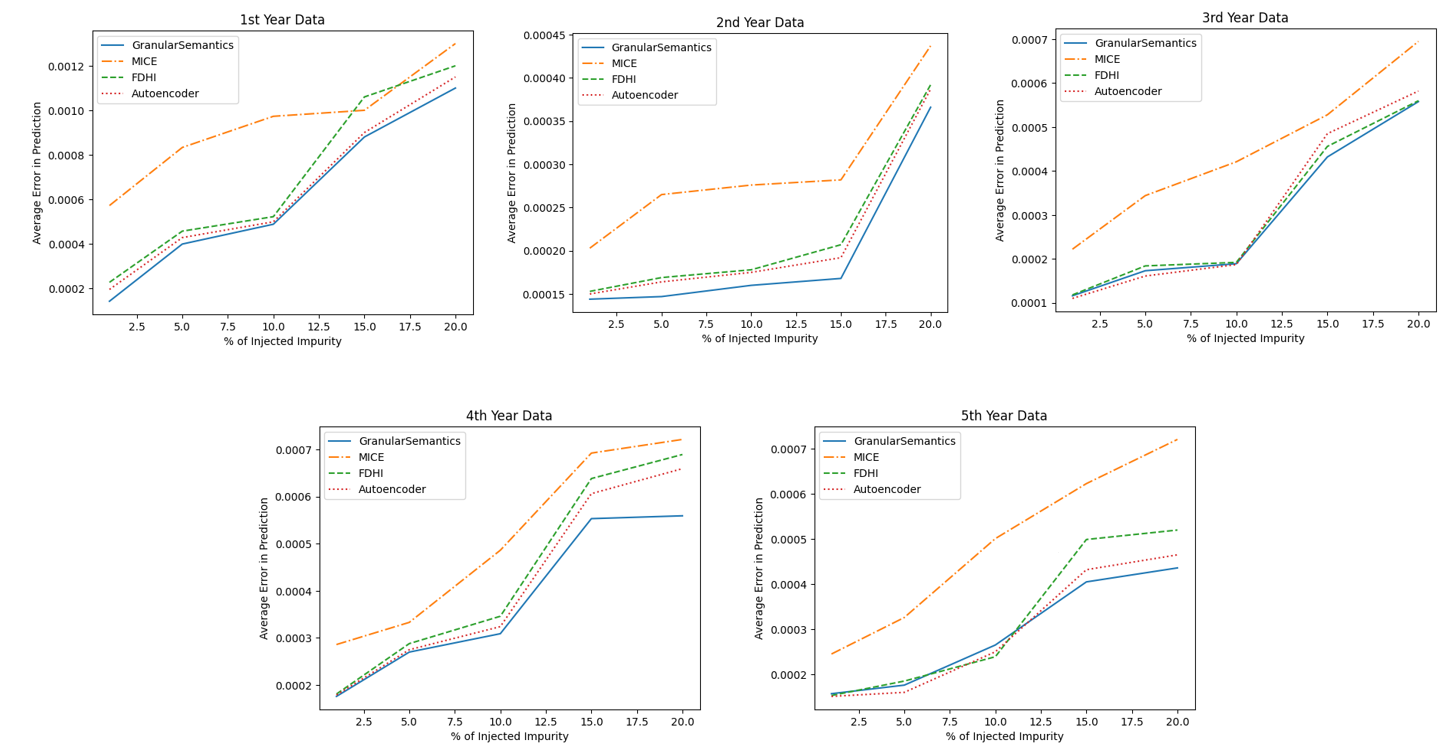}
    \caption{Variation in average error with increasing impurity over all the five year's data}
    \label{AE}
\end{figure}

To verify the reliability and robustness of the proposed method in comparison with the existing data imputation methods we performed the study by varying the amount of injected impurities. The results of it is shown in Fig. \ref{AE}. It can be observed from the figure that the performance of the proposed Granular Semantic method is almost consistent and as good as the other methods with low impurity ($<10\%$). Once the impurity increases, the proposed method performed better in all cases compared to the other methods. It proves the utility of considering the semantics of the feature and dropping the missing values while forming the granules. 

\subsection{Impact of Feature Reduction and Data Balancing}

\begin{figure}[h]
    \centering
    \includegraphics[scale=0.5]{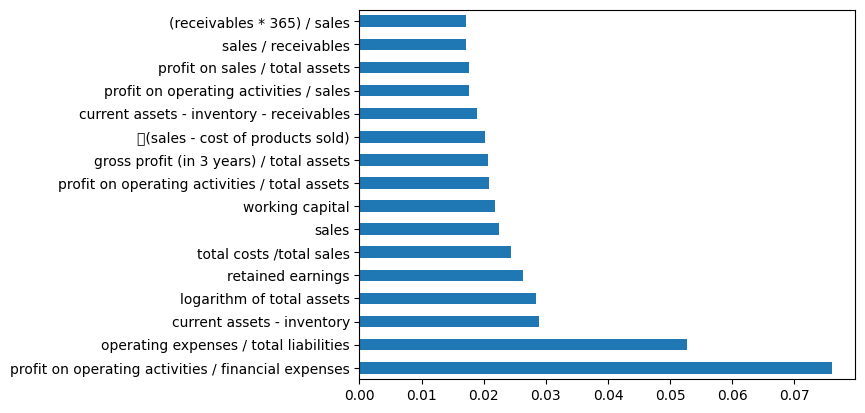}
    \caption{Features selected from Polish Bankruptcy Dataset for classificaition}
    \label{fs}
\end{figure}

In Polish bankruptcy dataset, 64 quantitative features are present there. Here random forest method has been used to select 16 most relevant features for bankruptcy prediction. As mentioned earlier, this work aims to design the entire model less computationally expensive to make it implacable for the small scale companies as well. The selected 16 features are shown in Fig. \ref{fs}.\\

Please note only the observations, without any missing values have been used here for this demonstration. Therefore, in this section $12789$ observations from the Polish Bankruptcy data set throughout all five years are used to demonstrate the effectiveness of feature reduction and data balancing in the pipeline. 

Most researchers, working on bankruptcy prediction, focus on large companies listed on the stock exchange platform, but small companies have only a limited number of attributes. Also, these small companies are not indexed on any stock exchange platform but cumulatively, they represent a significant part of the economy.  So to predict bankruptcy for such small companies, we trained the model with only 16 most important features. 

\begin{figure}[h]
\begin{subfigure}[b]{0.4\textwidth}
   \centering 
   \includegraphics[width=\textwidth]{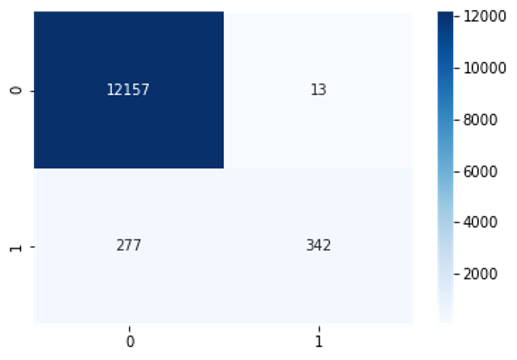}
    \caption{Confusion Matrix}
   \label{fig:nasa-logo}
\end{subfigure}%
\hfill
\begin{subfigure}[b]{0.4\textwidth}
   \centering 
   \includegraphics[width=\textwidth]{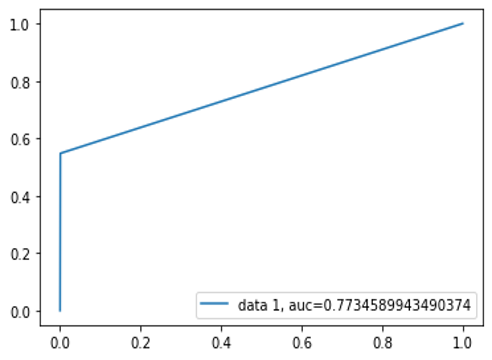}
    \caption{ROC Curve}
   \label{fig:artemis-logo}
\end{subfigure}%
   \caption{Result for Random Forest with  64 features} 
   \label{RF64}
\end{figure}

\begin{figure}[h]
\begin{subfigure}[b]{0.4\textwidth}
   \centering 
   \includegraphics[width=\textwidth]{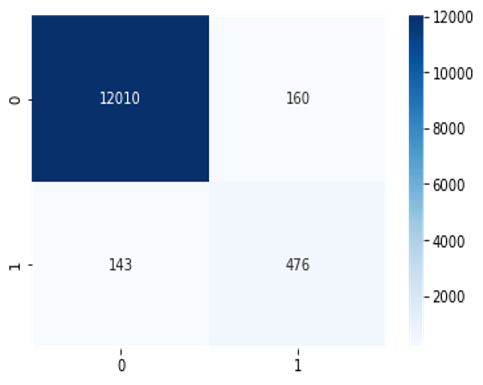}
    \caption{Confusion Matrix}
   \label{fig:nasa-logo}
\end{subfigure}%
\hfill
\begin{subfigure}[b]{0.4\textwidth}
   \centering 
   \includegraphics[width=\textwidth]{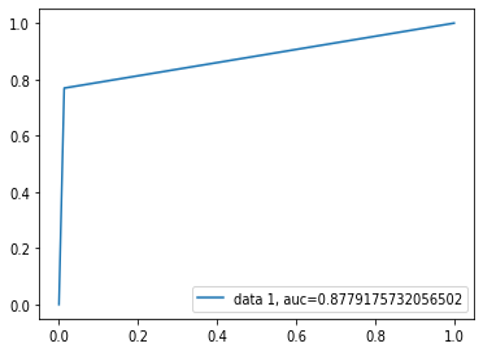}
    \caption{ROC Curve}
   \label{fig:artemis-logo}
\end{subfigure}%
   \caption{Result for Decision Tree with  64 features} 
   \label{DT64}
\end{figure}

\begin{figure}[h]
\begin{subfigure}[b]{0.4\textwidth}
   \centering 
   \includegraphics[width=\textwidth]{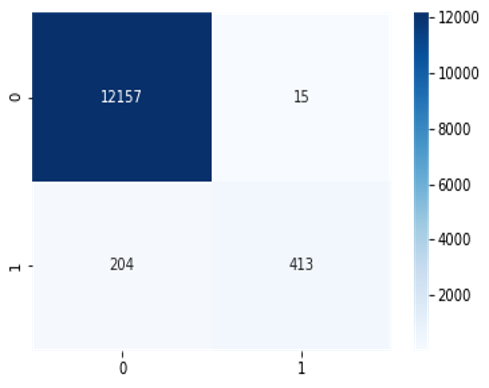}
    \caption{Confusion Matrix}
   \label{fig:nasa-logo}
\end{subfigure}%
\hfill
\begin{subfigure}[b]{0.4\textwidth}
   \centering 
   \includegraphics[width=\textwidth]{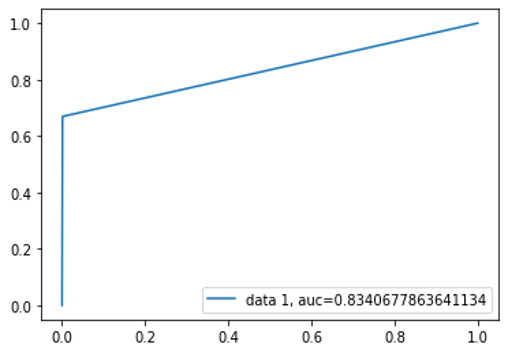}
    \caption{ROC Curve}
   \label{fig:artemis-logo}
\end{subfigure}%
   \caption{Result for Random Forest with  16 features} 
   \label{RF16}
\end{figure}

\begin{figure}[h]
\begin{subfigure}[b]{0.4\textwidth}
   \centering 
   \includegraphics[width=\textwidth]{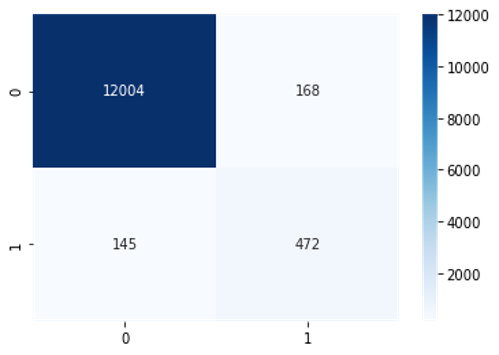}
    \caption{Confusion Matrix}
   \label{fig:nasa-logo}
\end{subfigure}%
\hfill
\begin{subfigure}[b]{0.4\textwidth}
   \centering 
   \includegraphics[width=\textwidth]{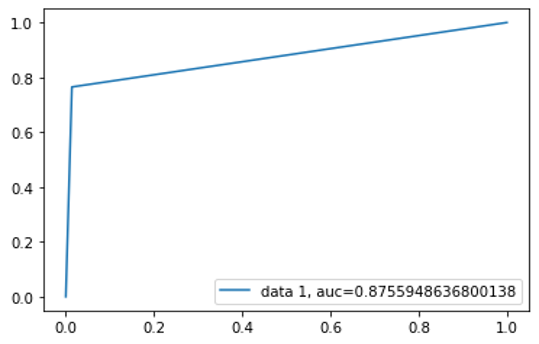}
    \caption{ROC Curve}
   \label{fig:artemis-logo}
\end{subfigure}%
   \caption{Result for Decision Tree with  16 features} 
   \label{DT16}
\end{figure}

\begin{figure}[h]
\begin{subfigure}[b]{0.4\textwidth}
   \centering 
   \includegraphics[width=\textwidth]{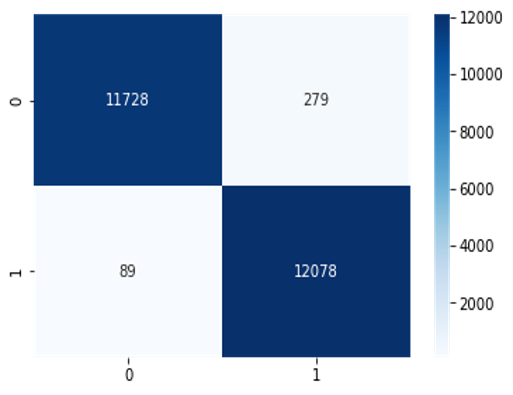}
    \caption{Confusion Matrix}
   \label{fig:nasa-logo}
\end{subfigure}%
\hfill
\begin{subfigure}[b]{0.4\textwidth}
   \centering 
   \includegraphics[width=\textwidth]{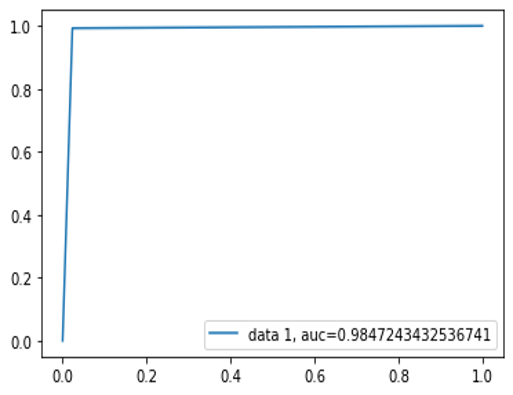}
    \caption{ROC Curve}
   \label{fig:artemis-logo}
\end{subfigure}%
   \caption{Result for Random Forest with 64 features+SMOTE} 
   \label{RF64S}
\end{figure}

\begin{figure}[h]
\begin{subfigure}[b]{0.4\textwidth}
   \centering 
   \includegraphics[width=\textwidth]{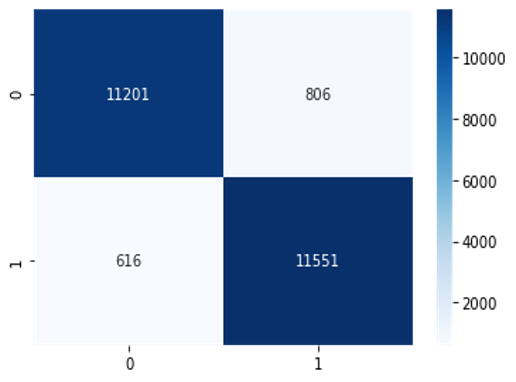}
    \caption{Confusion Matrix}
   \label{fig:nasa-logo}
\end{subfigure}%
\hfill
\begin{subfigure}[b]{0.4\textwidth}
   \centering 
   \includegraphics[width=\textwidth]{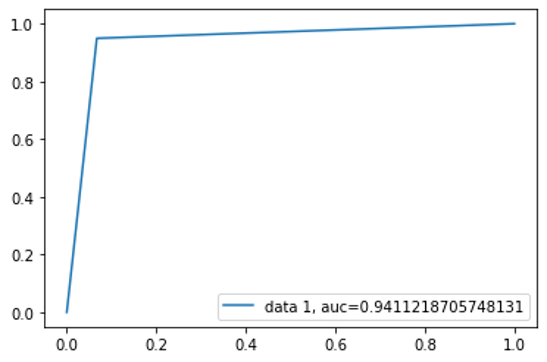}
    \caption{ROC Curve}
   \label{fig:artemis-logo}
\end{subfigure}%
   \caption{Result for Decision Tree with 64 features+SMOTE} 
   \label{DT64S}
\end{figure}

\begin{figure}[h]
\begin{subfigure}[b]{0.4\textwidth}
   \centering 
   \includegraphics[width=\textwidth]{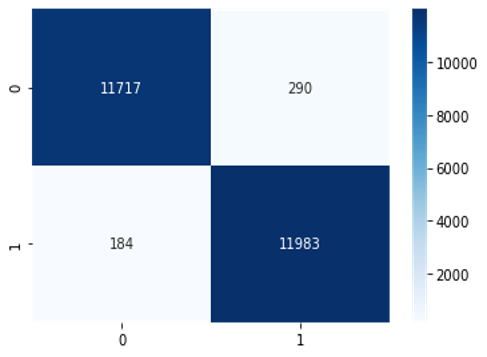}
    \caption{Confusion Matrix}
   \label{fig:nasa-logo}
\end{subfigure}%
\hfill
\begin{subfigure}[b]{0.4\textwidth}
   \centering 
   \includegraphics[width=\textwidth]{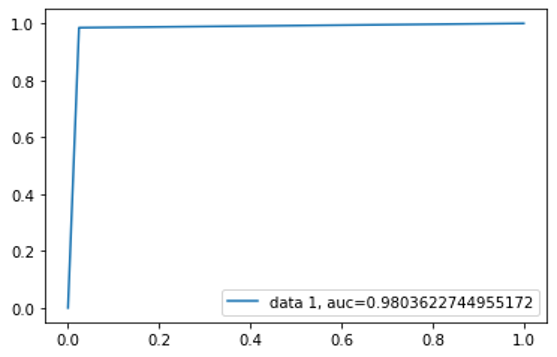}
    \caption{ROC Curve}
   \label{fig:artemis-logo}
\end{subfigure}%
   \caption{Result for Random Forest with 16 features+SMOTE} 
   \label{RF16S}
\end{figure}

\begin{figure}[h]
\begin{subfigure}[b]{0.4\textwidth}
   \centering 
   \includegraphics[width=\textwidth]{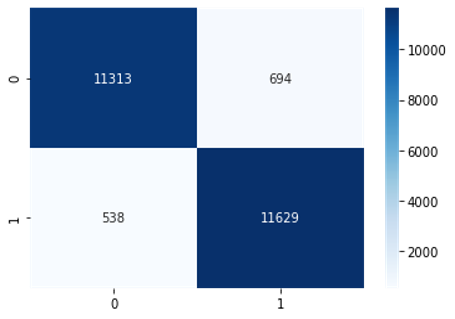}
    \caption{Confusion Matrix}
   \label{fig:nasa-logo}
\end{subfigure}%
\hfill
\begin{subfigure}[b]{0.4\textwidth}
   \centering 
   \includegraphics[width=\textwidth]{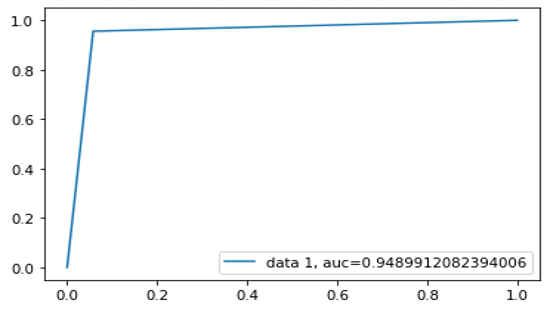}
    \caption{ROC Curve}
   \label{fig:artemis-logo}
\end{subfigure}%
   \caption{Result for Decision Tree with 16 features+SMOTE} 
   \label{DT16S}
\end{figure}

Since we are dealing with a highly imbalanced dataset here, where the number of bankrupt companies is much less than that of the non-bankrupted ones, the SMOTE \citep{chawla2002smote} method has been used to generate synthetic data in the minor class. The impact of feature reduction and data balancing over Polish bankruptcy dataset could be observed in the confusion matrices and ROC curves shown here from Fig. \ref{RF64} to Fig. \ref{DT16S}. The figures summarize the outcomes only for decision tree and random forest classifiers. As we can observe in Figs. \ref{RF64}(a), \ref{DT64}(a), \ref{RF16}(a), and \ref{DT16}(a), the models are working well with positive class, and have high true positive values, but fails in negative class prediction. That causes a high Type-2 error. This problem is well handled with class balancing of the dataset with SMOTE, and the results are reflected in Figs. \ref{RF64S}(a), \ref{DT64S}(a), \ref{RF16S}(a), and \ref{DT16S}(a) with a low Type-2 error. On the other hand, the impacts of proper feature selection are visible in the ROC curves. The ROC curves shown in Figs. \ref{RF64} (b), \ref{DT64} (b), \ref{RF64S} (b) and \ref{DT64S} (b) have lower AUC values compared to those of Figs. \ref{RF16} (b), \ref{DT16} (b), \ref{RF16S} (b) and \ref{DT16S} (b). It signifies that the reduction of redundant features causes a gain in accuracy. 

\subsection{Results of Bankruptcy Prediction}

This section demonstrates the effectiveness of the complete pipeline, that is, missing data filling with granular semantics, followed by feature reduction with random forest, and data balancing with SMOTE. The utility of the pipeline has been verified here for prediction of bankruptcy with the six different classifiers, as listed in Sec. \ref{ModelSel}. The results are summarized in Table \ref{Comparison}. Two metrics have been used to check the performance of the proposed method with the aforementioned six different classifiers. Those are accuracy and area under the curve (AUC). As it can be observed from Table \ref{Comparison}, the method defined in this work results in an accuracy around $90\%$ for all the dataset with all the six classifiers. The value of AUC is also around $0.8$ in all the cases, and it is as good as $0.9$ in some of them.   

\begin{table*}
\centering
    \caption{Bankruptcy Prediction with Proposed Pipeline Using Different Classifiers}
\small
\begin{tabular}{|c|c|c|c|c|c|c|c|c|c|c|}\hline\hline
   & \multicolumn{2}{|c|}{1st Year Data} & \multicolumn{2}{|c|}{2nd Year Data} & \multicolumn{2}{|c|}{3rd Year Data} & \multicolumn{2}{|c|}{4th Year Data} & \multicolumn{2}{|c|}{5th Year Data} \\\hline
     Classifier & Accuracy & AUC & Accuracy & AUC & Accuracy & AUC & Accuracy & AUC & Accuracy & AUC \\ \hline
     Logistic Regression & 0.921 & 0.827 & 0.899 & 0.852 & 0.932 & 0.853 & 0.941 & 0.862 & 0.875 & 0.811 \\ \hline
     K Nearest Neighbor & 0.895 & 0.803 & 0.834 & 0.766 & 0.901 & 0.841 & 0.892 & 0.831 & 0.913 & 0.824 \\ \hline
     Decision Tree Classifier & 0.881 & 0.814 & 0.872 & 0.817 & 0.925 & 0.882 & 0.913 & 0.854 & 0.875 & 0.810 \\ \hline
     Random Forrest Classifier & 0.934 & 0.837 & 0.951 & 0.862 & 0.927 & 0.873 & 0.944 & 0.882 & 0.929 & 0.842 \\ \hline
Gradient Boosting & 0.876 & 0.792 & 0.855 & 0.781 & 0.923 & 0.823 & 0.902 & 0.850 & 0.890 & 0.801 \\ \hline
Deep Neural Network & 0.952 & 0.881 & 0.940 & 0.893 & 0.948 & 0.836 & 0.939 & 0.891 & 0.938 & 0.861 \\ \hline
\end{tabular}\label{Comparison}
\end{table*}
Two metrics have been used to check the performance of the proposed method with the six different classifiers mentioned above. Those are the accuracy and the area under the curve (AUC). As it can be observed from Table \ref{Comparison}, the method defined in this work results in an accuracy around $90\%$ for all the dataset with all the six classifiers. The value of AUC is also around $0.8$ in all cases and is as good as $0.9$ in some of them.

\section{Conclusions and Discussions}\label{con}
The overall method defined here for bankruptcy prediction has been proven to be effective over all the five years Polish dataset. The newly formulated data imputation technique with contextual granule has been compared with three other popular methods, and resulted in higher or almost equal accuracy even compared to autoencoder-based estimators. Moreover, this imputation method has reflected its robustness while tested with the increasing rate of missing values, and henceforth it has proven its reliability. The effectiveness of the entire pipeline has also been demonstrated with the impacts of feature reduction and data balancing. The end-to-end pipeline designed here results in accuracies more than $90\%$ for the prediction of bankruptcy in most cases. However, the proposed data imputation method could be verified with other high-dimensional datasets, and its prediction accuracy with categorical data could be checked. This imputation method may not be much efficient once the impurity is more than $50\%$, since more than half of the database may need to be scanned while forming the granules around each missing entry, thereby making it computationally rigorous. Further, the pipeline designed here could also be validated with other bankruptcy datasets.
\bibliographystyle{unsrtnat}
\bibliography{refs}  






\end{document}